%% file: main.tex
\documentclass[journal]{IEEEtran}
\usepackage[style=ieee,backend=biber]{biblatex} % biblatex
\usepackage{hyperref}

\ifCLASSINFOpdf
   \usepackage[pdftex]{graphicx}
    \graphicspath{ {./src/img/} }

\else
\fi
\usepackage{tabularx}
\newenvironment{conditions*}
  {\par\vspace{\abovedisplayskip}\noindent
   \tabularx{\columnwidth}{>{$}l<{$} @{${}={}$} >{\raggedright\arraybackslash}X}}
  {\endtabularx\par\vspace{\belowdisplayskip}}

\usepackage{amssymb}
\usepackage{mathtools}
\usepackage{atbegshi}% http://ctan.org/pkg/atbegshi
\AtBeginDocument{\AtBeginShipoutNext{\AtBeginShipoutDiscard}}
\begin{document}
\title{Fully Homomorphically Encrypted\\ Deep Learning as a Service}
%\author{Anonymous}% <-this % stops a space

\author{George~Onoufriou,
       Paul~Mayfield
        and~Georgios~Leontidis}% <-this % stops a space
\thanks{G. Onoufriou was with the Department
of Computing Science, University of Aberdeen, Aberdeen
AB243UE UK and now is with the School of Computer Science, University of Lincoln, Lincoln, LN67TS, UK, email:gonoufriou@lincoln.ac.uk}% <-this % stops a space
\thanks{P. Mayfield is with the Scotland's Rural College and SAC Consulting, email: Paul.Mayfield@sac.co.uk}% <-this % stops a space
\thanks{G. Leontidis is with the Department of Computing Science, University of Aberdeen, Aberdeen, AB243UE, UK, email:georgios.leontidis@abdn.ac.uk} 

\maketitle

\begin{abstract}
  \input{src/sections/abstract.tex}

\end{abstract}


% \fi
\IEEEpeerreviewmaketitle

\section{Introduction}
  \input{src/sections/introduction.tex}

\section{Fully Homomorphic Encryption}
  \input{src/sections/fhe.tex}
  \nocite{*}

\section{Methodology}
\input{src/sections/methodology.tex}

\section{Results and Discussion}
\input{src/sections/results.tex}

\section{Conclusion}
\input{src/sections/conclusion.tex}

% or


\appendices

% \section{}

\section*{Acknowledgment}

This project received funding from the UKRI-EPSRC grant "The Internet of
Food Things" (EP/R045127/1).

\ifCLASSOPTIONcaptionsoff
  \newpage
\fi



%\bibliographystyle{IEEEtran}
% \end{thebibliography}

% % using biblatex version mimicking their style
\printbibliography


%\end{IEEEbiography}


%\newpage



%\vfill


\end{document}

%% file: src/sections/abstract.tex
Fully Homomorphic Encryption (FHE) is a relatively recent advancement in the field of privacy-preserving technologies. FHE allows for the arbitrary depth computation of both addition and multiplication, and thus the application of abelian/polynomial equations, like those found in deep learning algorithms.
This project investigates, derives, and proves how FHE with deep learning can be used at scale, with a relatively low time complexity, the problems that such a system incurs, and mitigations/solutions for such problems. In addition, we discuss how this could have an impact on the future of data privacy and how it can enable data sharing across various actors in the agri-food supply chain, hence allowing the development of machine learning-based systems.
Finally, we find that although FHE incurs a high spatial complexity cost, the time complexity is within expected reasonable bounds, while allowing for absolutely private predictions to be made, in our case for milk yield prediction.

%% file: src/sections/introduction.tex
\IEEEPARstart{T}{rust}, data quality, data quantity, and data integrity are critical ingredients required for the successful application of deep learning:

\begin{itemize}
    \item Trust; is necessary to access data in the first instance -- without trust there is usually a rightful unwillingness to collaborate and subsequently share data, unless the data/ collaboration is very insensitive / cannot be harmful, or purposefully open for some other reason.
    \item Data Quality; good quality data is necessary to train, and infer, if it does not have internal consistency / regularity and relation to some output then it can be difficult or impossible to seek to use this data for prediction of this output.
    \item Data Quantity; there is some threshold of which anything less than this amount of data even under perfect use simply does not hold enough information to properly train or to be able to make a reasonable inference.
    \item Data Integrity; both that there exists a lack of tampering, and that the data is properly representative of the scenario it tries to predict are important. The lack of the former would leave room for potential malicious actors to disrupt predictive algorithms, toward some damage or harm. The latter would make any successfully trained neural network useless in real scenarios, because the distribution it has learned is not properly representative of the ground truth.
\end{itemize}

We seek to help solve some of these problems, as far as is reasonably possible, by the use, application, and evaluation of fully homomorophically encrypted deep learning at scale:

\begin{itemize}
    \item Trust; if there is no requirement for trust, being that the data is undecryptable, and the whole process is an auditable open kerckoffian procedure in both encryption and processing, then this barrier to collaboration is removed or at-least significantly mitigated.
    \item Data Quality; unfortunately encrypting data in this manner before it can be analyzed and properly engineered by a potentially more specialized entity will naturally mean the data quality cannot be assured during processing, since there is no means by which to verify it. However given that other barriers are being diminished it is possible that highly detailed sensitive data could now become available.
    \item Data Quantity; data that would otherwise be too litigatious or sensitive to use could now become readily available in a fully homomorphically encrypted form, significantly increasing the quantity of usable data toward better trained although single purpose/ bespoke models.
    \item Data Integrity;  similarly to data quality, the relevance of the data to the underlying ground truth cannot be assessed, but there is some protection towards tampering, being that any encrypted cyphertext that did not originate from the data owner is not decryptable by that owners keys, nor processable by their public, reliniarisation, etc keys, halting the process immediately.
\end{itemize}

Especially if this can be accomplished at some scale, we believe reducing these barriers by implementing fully homomorphically encrypted deep learning as a service (EDLaaS) represents a large stride towards a more private, and secure transaction between data owners and data processors. EDLaaS would also mean a large change in the development, training, and general principles of deep learning towards a much more sustainable future for this science, in which privacy requirements are satisfied, and fears from risk are significantly reduced if not nullified.

An example, where EDLaaS could be a benefit to individual users, is that of home assistants, such as Google home, and Alexa. Given that the end devices could each individually house their own private keys, user voice could be fully homomorphically encrypted, and still used by the back-end service to divine the instructions, which are to be executed by the same end device; thus protecting the consumer from data privacy worries, and the operator from GDPR \autocite{gdpr} concerns in both transit and operational-use, since this never requires decryption outside of the home or is indeed indecryptable by the back-end system.

Another example, however this time for agriculture/ industry, in which we have actively explored and will discuss it further as part of this paper, is that of a data processor for milk yield prediction. In many industries there are large concerns over potential data leaks over perceived or real sensitivity latent in their data such as genetics / breeding, or feed composition in the case of the milk industry. In addition, concerns around food traceability and safety, along with how information can be safeguarded against malicious input are very important in the agri-food sector \autocite{pearson2019distributed}. Moreover, considering that data in such industries is a very valuable intellectual property, stakeholders are hesitant to share their data, even when they are seeing some benefits in doing so. There is too much at stake for them, therefore solutions that could enable data sharing or alternatively sharing encrypted data that can be used to develop machine learning applications, would be a game changer for the sector \autocite{durrant2021might}. To test and evaluate our implementation we were provided with the last 30 years of breeding, feeding, and milk yields data, by the Langhill Dairy herd based at the SRUC Dairy Innovation Centre, Dumfries.

For the sake of clarity, FHE \autocite{gentry2009fully} or more specifically the CKKS \autocite{cheon2017homomorphic, cheon2018bootstrapping} implementation already exist as a technology, therefore we do not seek to prove that it is secure even though we believe that it currently is; instead our contribution is the derivation of a method along with an application that show how it can be used not just in laboratory conditions but also in production like environments to function as a means to conduct encrypted deep learning as a service, the penalties we incur when adopting such a technology, and our solutions to other problems along the way. This way we can help bring FHE out of emerging/ proof of concept status, by doing much of the hard work needed to use it at scale.

\subsection{Motivation}
Neural networks (NNs) are an ML algorithm that can be used in various settings and with many types of data, from images and time series to cloud points and fourier-transformed data. NNs in the context of agri-food have already been used across a number of settings, e.g. yield forecasting \autocite{alhnaity2019using, alhnaity2021autoencoder, durrant2021role}, crop and fruit detection \autocite{hossain2018automatic}, pest detection \autocite{cheng2017pest}, etc. To train (deep) neural networks and exploit their full potential we require more and more data. There are concerns on behalf of the data owners, specifically on the sensitivity of their data, and its (mis)use. Their sensitivity creates a reluctance to share, especially if the collaboration is new, as there is a lack of trust and issues around background and foreground IP. Thus if we want to create more and new collaborations in order to enable net-zero transition, enhance environmental sustainability and improve productivity, it is necessary to build up this trust or create a system, where they do not need to trust the data processor as it will work with encrypted data. This lends itself to fully homomorphically encrypted data, as they no longer need to trust the data processor, their data cannot be decrypted, read, or leaked, but it can still be used for computation to produce effective predictions for the data owners.

%A recent report from the Royal Society (Royal Society) highlighted the importance of enabling trust via the establishment and adoption of PPE technologies, suggesting that more investment is needed to drive innovation in this domain.

%In addition EPSRC have identified some themes that are driving their strategies, one of which is information and communication technologies, under which a number of key research areas relate to our project, such as artificial intelligence technologies, and also privacy preserving technologies, trust and cybersecurity.

Stemming from all the above our motivation has been to investigate how FHE can be used to enable the exchange of encrypted data in an agri-food setting and enable the use of ML as part of the pipeline via the use of edge devices and virtual machines (VMs). FHE as part of an ML pipeline is still in its infancy \autocite{privacy2019royal}, which means that in contrast with ML methods that work with non-encrypted data, no off-the-shelf approaches exist that can be routinely used, making their adoption a hard process, hence still considered an emerging and possibly disruptive technology.

Finally, the aim of this paper is to test the feasibility of using FHE on dairy milk data. In simple terms, applying FHE to a set of data enables useful operations to be conducted on encrypted values without decrypting them first. Both the input and output data remain encrypted (see figure 1). Homomorphic encryption solves a vulnerability inherent in all other approaches to data protection. Currently, traditional public key encryption requires that data be decrypted before it can be analyzed or manipulated, exposing the data to security and manipulation threats while in the decrypted state. FHE allows any data to remain encrypted while it’s being processed, eliminating the need for decryption and providing another layer of security to the data. Such a tool can be used to develop trust in data sharing between businesses, practitioners and research organisations in the food and drink sector.

%% file: src/sections/fhe.tex
Fully Homomorphic Encryption (FHE) is a structure-preserving encryption transformation \autocite{gilad2016cryptonets} first appearing in 2009 \autocite{gentry2009fully}, and having several advancements since to improve its efficiency, and speed. \autocite{gilad2016cryptonets} FHE largely depends on commutative algebra, in particular modeling the ring learning with errors (RLWE) problem. Commutative rings are sets in which it is possible to add, subtract (via the additive inverse), and multiply, and still result in a member of the set \autocite{gilad2016cryptonets} . For more in depth detail about rings, fields and the associated axioms that must be met by any deep learning algorithm please see \ref{subsec:rings}
However, in summary the primary consequence to not being a field is the lack of divisibility, since we do not have access to the multiplicative-inverse, whereas a field can always guarantee the additive inverse, meaning we can still subtract by addition of a negative. The lack of division will undoubtedly cause issue with things such as activation functions if we were to use a sigmoid function ($\frac{1}{1+e^{-x}}$) meaning we should use a different function or approximate. One such possibility is a Taylor expansion series which closely approximates sigmoid, although there are a number of alternative methods proposed \autocite{bos2014private, song2019bitwise}. \\

FHE/RLWE has recently been paired with deep learning with success, causing some movement toward FHE in literature but primarily for convolutional neural networks (CNNs). There is a gap in that homomorphic encryption has not been applied to more complex and up to date methods, nor real world problems with their own levels of added considerations, which is reinforced by the aforementioned Royal Societies statement that FHE is still a proof of concept technology \autocite{privacy2019royal}. The majority of research in privacy-preserving neural networks is with CNNs applied to the MNIST dataset \autocite{nandakumar2019towards, riazi2019deep, boura2018chimera}. The key aspect focused in these papers is the need to use different FHE compatible equivalents for things such as the activation functions, e.g. sigmoid approximation. This is because depending on the FHE implementation, only addition and multiplication may be applied to the cipher-text and still maintain its usability. In these same papers they suggest that FHE is 4-5 times slower to train and infer than the unencrypted data, so this is also something that should be sought for improvement.

\subsection{Commutative Rings Formalisation}
\label{subsec:rings}

Commutative rings are sets in which it is possible to add, subtract (via the additive inverse), and multiply, and still result in a member of the set. This includes the sets:
\begin{itemize}
  \item $\mathbb{Z}$; integers, E.G.: $(-1, 0, 1, 2, ...)$ Formally: An integer is any number that has no fractional part (not a decimal).
  \item $\mathbb{Q}$; rational numbers, E.G.: $(5, 1.75, 0.001, -0.1, ...) = (\frac{5}{1}, \frac{7}{4}, \frac{1}{1000}, \frac{-1}{10}, ...)$
  Formally; a rational number is a number that can be in the fractional form $\frac{a}{b}$ where $a$ and $b$ are integers and b is non-zero.
  \item $\mathbb{R}$; real numbers, E.G.: $(0, -1.5, 3/7, 0.32, \pi)$ Formally; a real number is any non-imaginary, non-infinite number.
  \item $\mathbb{C}$; complex numbers, E.G.: $(1+i, 32+-2.2i, 5, -6i)$ Formally: A number which is a combination of real and imaginary numbers, where either part can be zero.
\end{itemize}
This does not include the sets:
\begin{itemize}
  \item $\mathbb{I}$; imaginary numbers, E.G $ where: i=\sqrt{-1}, (i, -i, 39.8i, ...)$ Formally: Imaginary numbers are any numbers which are multiplied by the imaginary unit $i$.
\end{itemize}
$R$ for (commutative) ring shall henceforth be one of the four sets $\mathbb{Z}, \mathbb{Q}, \mathbb{R}, \mathbb{C}$. In contrast a \textit{field (F)} is any \textit{commutative ring ($R$)} which may also perform division and still result in elements from that ring. This includes only the sets $\mathbb{Q}, \mathbb{R}, \mathbb{C}$ as not all elements in the set of integers ($\mathbb{Z}$) can be divided by another integer and still result in an integer \autocite{ershov2015lec}.
These rings are used through polynomial expressions instead of discreet matrices in the learning with errors (LWE) thus ring learning with errors (RLWE). For formalization if all of the following axioms are fulfilled then the resulting set is called a field:\\
$addition~axioms;$ \\
$given: (x,y,z \in R), then:$
\begin{equation}
  \label{equation:axioms_addition}
  \begin{aligned}
    (unity)~0 \in R \\
    (closed)~x+y \in R \\
    (inverse)~x,-x \in R \\
    (commutative)~x+y = y+x \\
    (associative)~(x+y)+z = x+(y+z) \\
  \end{aligned}
\end{equation}
$multiplication~axioms;$ \\
$given: (x,y,z \in R), then:$
\begin{equation}
  \label{equation:axioms_multiplication}
  \begin{aligned}
    (unity)~1 \in R \\
    (closed)~x\cdot y \in R \\
    (inverse)~x,x^{-1} \in R \\
    (commutative)~x\cdot y = y\cdot x \\
    (associative)~(x\cdot y)\cdot z = x\cdot (y\cdot z) \\
  \end{aligned}
\end{equation}
$multiplicative~additive~axioms;$ \\
$given: (x,y,z \in R), then:$
\begin{equation}
  \label{equation:axioms_both}
  \begin{aligned}
    (distributivity)~(x+y)\cdot z=x\cdot z+y\cdot z
  \end{aligned}
\end{equation}
If all but multiplicative-inverse then this is a commutative ring with 1, if this also does not fulfil multiplicative-unity then this is just a commutative ring \autocite{ershov2015lec}.

%% file: src/sections/methodology.tex
\begin{figure}[th!]
  \centering
  \includegraphics[width=\columnwidth]{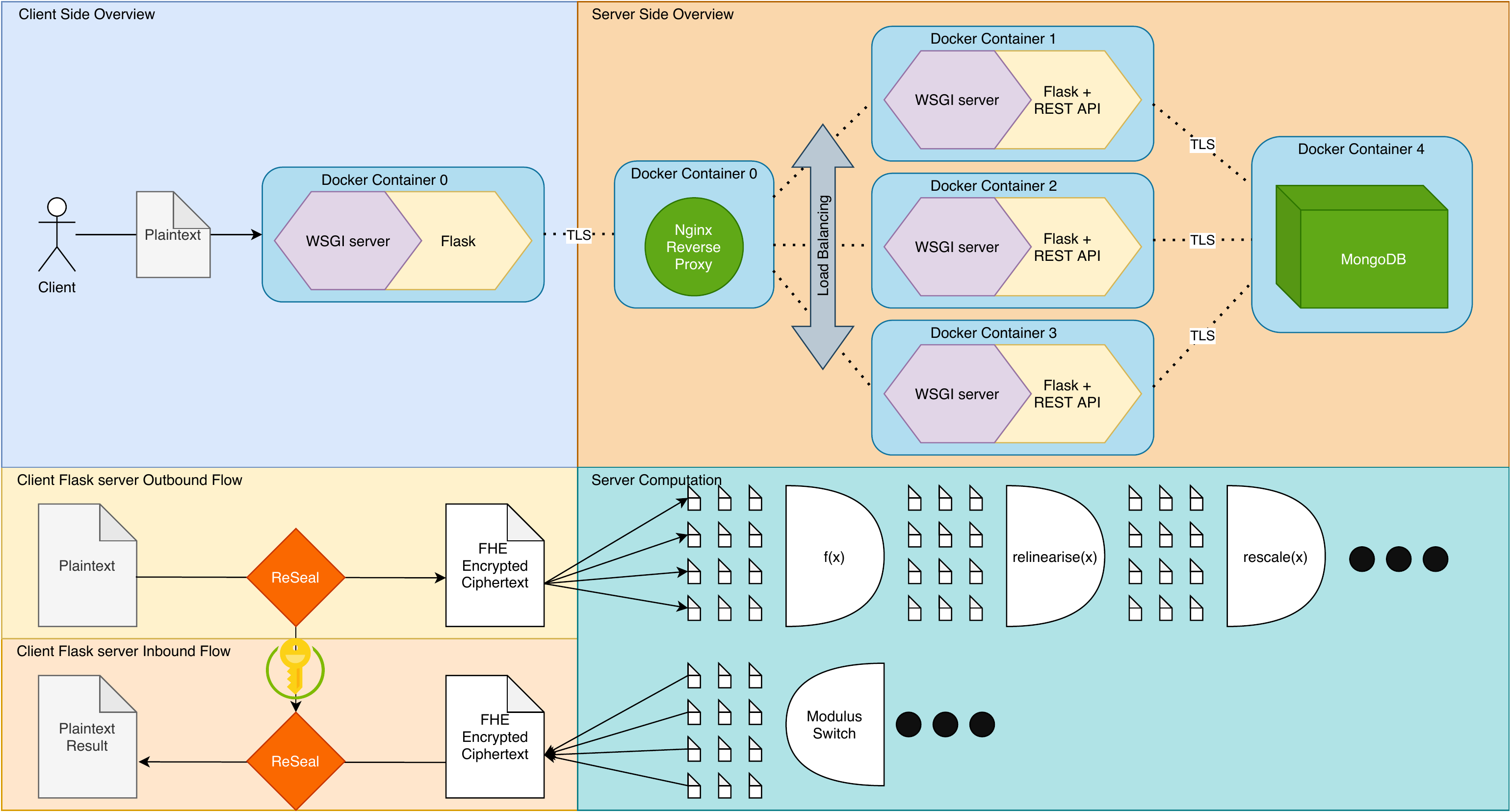}
  \caption{Pipeline that demonstrates the key stages of our project’s pipeline, from the client and raw data (upper left) to the data processing and analytics (lower right).}
  \label{fig:pipeline}
\end{figure}

To evaluate FHEs applicability as an EDLaaS we needed to create and mimic as closely as possible what we expect to be a standard industrial use case for third party data processing, which we can evaluate the effect of FHE on, along with evaluating FHE's time and spatial complexity itself. To this end we devised a two part client server system depicted in Figure \ref{fig:pipeline}.

Towards the end of creating this evaluable pipeline we had to overcome a few shortfalls we found at the time that prevented FHE to be integrated into an EDLaaS scenario/ pipeline.

\begin{itemize}
    \item Fully Homomorphic encryption itself, specifically CKKS \autocite{cheon2017homomorphic, cheon2018bootstrapping}, and adapting it to be usable at some scale.
    \item Combining FHE with deep learning, which had only been peripherally explored at this point.
\end{itemize}

\subsection{Data Pipeline}

Broadly our data pipeline can be abstracted into a few different categories, necessary to test and evaluate FHE at some scale, and in a practical manner to garner real results; Data source where data is wrangled and encrypted, and data sink where the data is processed on much more powerful and fully featured machines.

\subsubsection{Data Wrangling}

For our study we used data on dairy herds over the last 30 years provided by the Langhill Dairy herd based at the SRUC Dairy Innovation Centre Dumfries. We use this data as ground truth with which to encrypt and infer on using time series neural networks, in our case a one dimensional convolutional neural network (1D CNN). We normalised this data between the range of 0-1, one hot encoded categorical features, and used the historic feeding, genetics, and subsequent milk yield as examples in time leading up to the current milk yield prediction. This is fundamentally the same as any other data wrangling where data is prepared for processing by neural network, with the sole exception that the data is then encrypted, meaning this is the final form of the data, and cannot be changed before training, but can of course be iteratively adapted if it does not provide the best results by simply feeding more but differently wrangled encrypted data. Since there is usually a significant number of empty slots in the encryption vector, it may be possible to optimize further by merging multiple examples into a single encrypted vector. However, this emptiness allows for a lot of variance in wrangling techniques without the need to create whole new neural network architectures.

\subsubsection{Client/ Data Source}

The data source, usually a small embedded device (in our case, NVIDIA jetson nano), is responsible for data wrangling and encryption, since once encrypted the data can no longer be seen, and cannot be verified; thus the need to transform pre-encryption. The data source must be the encryptor so that they are the only entity with a private key with which to decrypt the data again.

Normally, the data owner cannot be expected to be familiar with FHE, deep learning, and thus the requirements of the data to be properly processable. It is necessary that some form of interaction/ awareness of the data occurs such that appropriate auditable/ open-source data processors can be provided. In the ideal scenario this would not be necessary and the data owner would be capable of wrangling the data according to their needs, but it should be noted this is an unlikely occurrence given there is more data sources than expertise, and the existence of expertise reduces the likelihood in the need for an EDLaaS. However given client expertise or at-least proficiency in data cleaning, and use of open-source helpers and documentation, then no embedded device would be necessary, and the client can instead submit their cyphertext directly for processing.

Data from the client is serialised by the embedded device after encryption ready for transmission to the data processor, without the presence of the private key, ensuring the transmitted serialised cyphertext is undecryptable during processing in the later stages. These keys should instead remain indexed on the client machine/ embedded device.

\subsubsection{Server/ Data Processor}

Data from the data source is serialised and transmitted using standard https requests, to model how it would likely function in such an Internet service. The data sink then proceeds to deserialise and apply the arbitrary computation, in our case a neural network, and then serialises and transmits the still encrypted but now transformed data to the data source for final decryption and use. An example encrypted output can be seen in the following figure \ref{fig:encrypted_data} showing cyphertexts, the associated keys, the data set name that it is associated with, who owns this data, and when it was submitted. In practice, we would filter unnecessary information, like the plain parameters, to reduce space and time costs of storing and transmitting this data, hence improving speed; and of course we would not handle the private key, which is shown here for experimental purposes only.

We can process data as in figure \ref{fig:encrypted_data} using our library described in \ref{subsec:fhe_library} and with our techniques outlined in \ref{subsec:encrypted_deep_learning}. Once processed to minimise space consumption, we swap all the way down the remaining coefficient chain, to create the smallest possible, and most quickly deciphered cyphertext, thus saving space and time while the data is stored until the data owner decrypts/ uses the results.

\begin{figure}[th!]
  \centering
  \includegraphics[width=0.8\columnwidth]{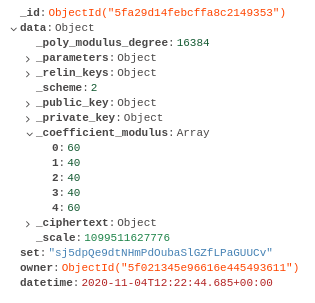}
  \caption{Serialised representation of encrypted data using CKKS scheme, and including all private, relin, and public keys, where objects here are byte arrays.}
  \label{fig:encrypted_data}
\end{figure}

\subsection{Interface}

\begin{figure}[th!]
  \centering
  \includegraphics[width=\columnwidth]{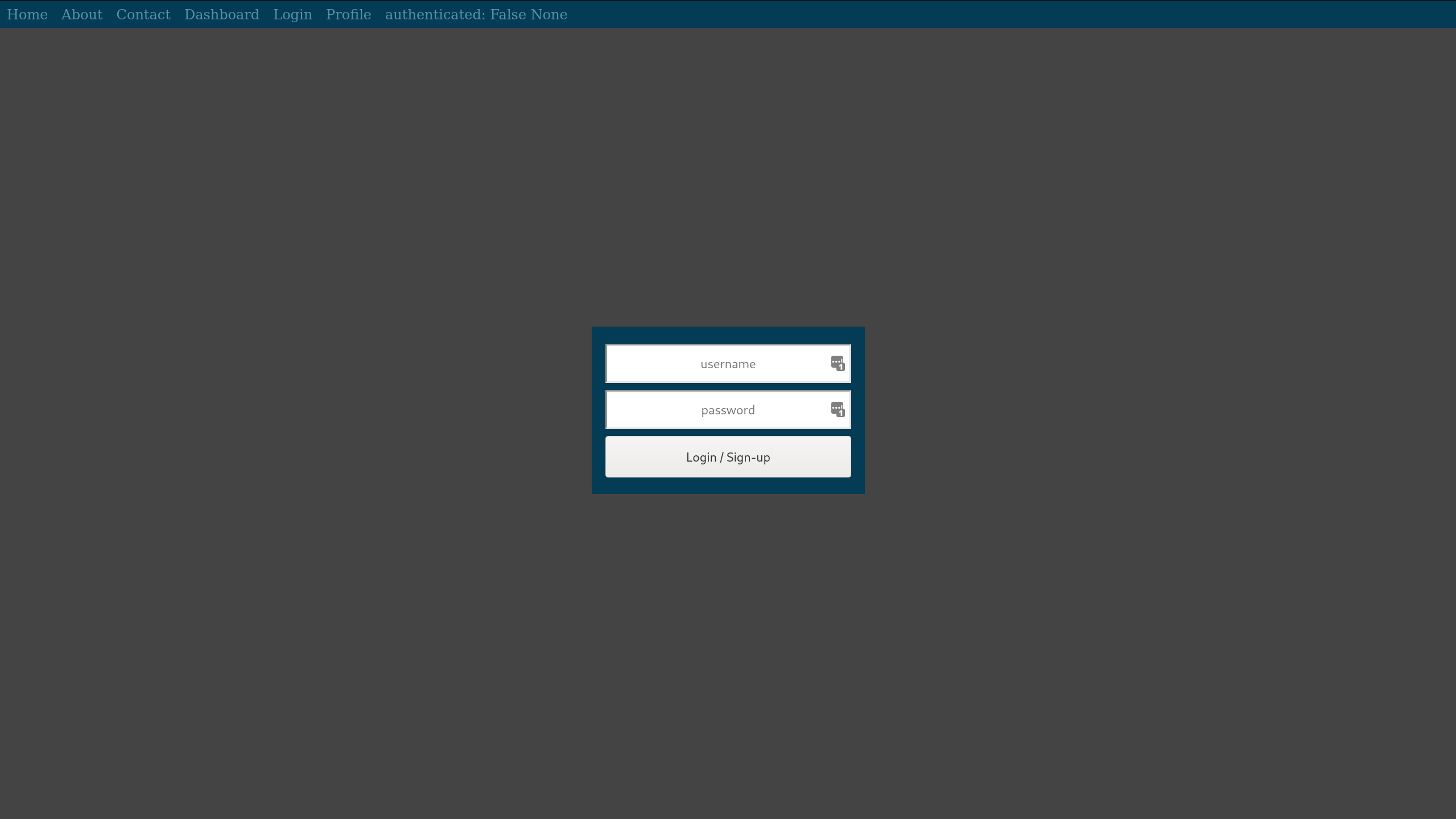}
  \caption{Authentication of local and/ or remote interface, avaliable at nextgen.abdn.ac.uk}
  \label{fig:login}
\end{figure}

\begin{figure}[th!]
  \centering
  \includegraphics[width=\columnwidth]{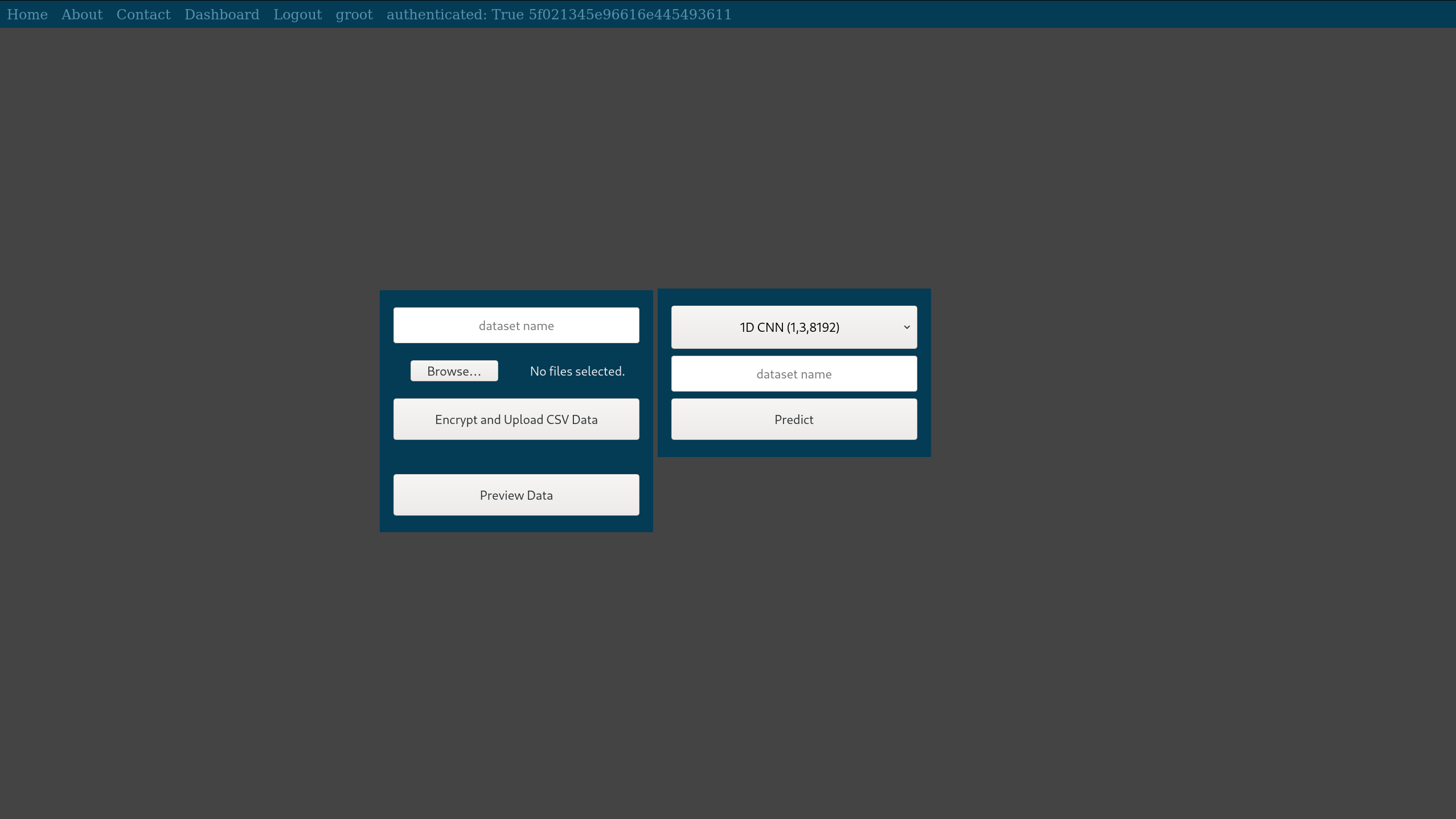}
  \caption{FHE dashboard, allowing simple upload, data view (of metadata since data is encrypted), and processing of data.}
  \label{fig:dash}
\end{figure}

\begin{figure}[th!]
  \centering
  \includegraphics[width=\columnwidth]{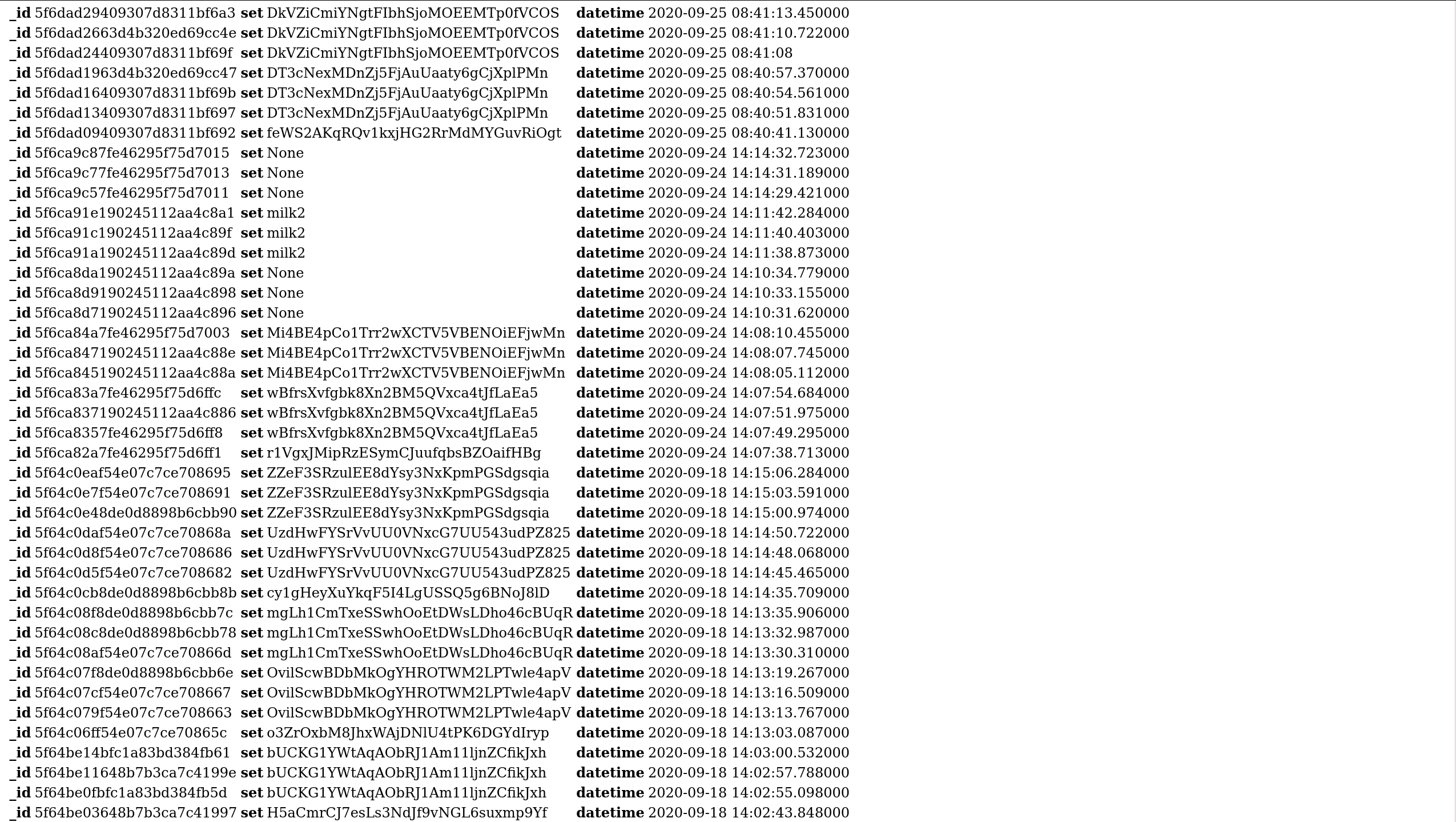}
  \caption{Minimal data preview, so users can assess what data exists, by date and given data set name for reference.}
  \label{fig:preview}
\end{figure}

For ease of use and testing we created a web app interface to be used by the data owner to simplify the process of submitting data to the embedded client device for encryption and transformation, while also serving the purpose of user authentication (figure \ref{fig:login}), as a main dashboard that prompts the user to upload the data to be encrypted (figure \ref{fig:dash}), and also an opportunity to view the encrypted data (figure \ref{fig:preview}). The training process, given the complexity involved, can run either on the jetson device or a remote host, e.g. high-performance computing. The user may also select what type of 1D CNN to use to provide them predictions. As far as our techniques are concerned, they can run in both an edge and a non-edge device, depending on the scalability of the problem and other constraints related to the amounts of data, training time, etc.

\subsection{Fully Homomorphic Encryption Library}
\label{subsec:fhe_library}

 At the time in the community there was little work available to easily use fully homomorphic encryption in conjunction with deep learning frameworks, and of what was available they often did not support the more complicated Cheon, Kim, Kim and Song (CKKS) \autocite{cheon2017homomorphic, cheon2018bootstrapping} FHE scheme, and its serialisation, or harbor some hidden catches. The CKKS scheme can operate on floating precision numbers which is critically necessary since normalised neural networks operate on floating point numbers usually in the range 0-1. The base library we used after quite some deliberation was the Microsoft Simple Encrypted Arithmetic Library (MS-SEAL) since it supports CKKS and some form of serialisation necessary to broadcast over the internet. However this did not solve the incompatibility with both GPU compute and deep learning libraries which MS-SEAL did not have. As such we created our own open-source abstraction library that would allow us to use python and python libraries to speed up our research and development of our system, at some scale along with using it in conjunction with web servers easily. Secondly MS-SEAL does not support bootstrapping yet, meaning there is a limit to the number of computations we can process, but since we were intending to stay within this limit for the sake of noise budgeting.

 Most deep learning applications, and the associated plethora of libraries primarily use the programming language python. MS-SEAL is written in C++, C\#, but not python. Thus if we intend to use MS-SEAL in conjunction with other machine learning libraries it was a requirement to create bindings from the C++ implementation to python. Luckily there existed many early attempts to create MS-SEAL python bindings, most of which were using already outdated versions of MS-SEAL that did not have good serialisation support. We found a few likely candidates in the community that were using current versions which we collaborated with and extended their implementation to form our first basic serialisable version of MS-SEAL bound to python. \autocite{python-seal} These bindings however would need some further abstraction in python to make them usable at scale, which lead to a further effort creating our ReSeal library, which includes all the serialisation logic, and abstractions necessary to make MS-SEAL easily usable in conjunction with larger frameworks. Our ReSeal implementation is open source (OSLv3 Licenced), and freely available. \autocite{python-reseal}

 We would like to in future, improve on some aspects of serialisation, and to make installation of ReSeal easier, and this is something we intend to iterate, and improve on over time.

\subsection{Fully Homomorphic Encryption in Deep Learning}
\label{subsec:encrypted_deep_learning}

\begin{figure}[th!]
  \centering
  \includegraphics[width=\columnwidth]{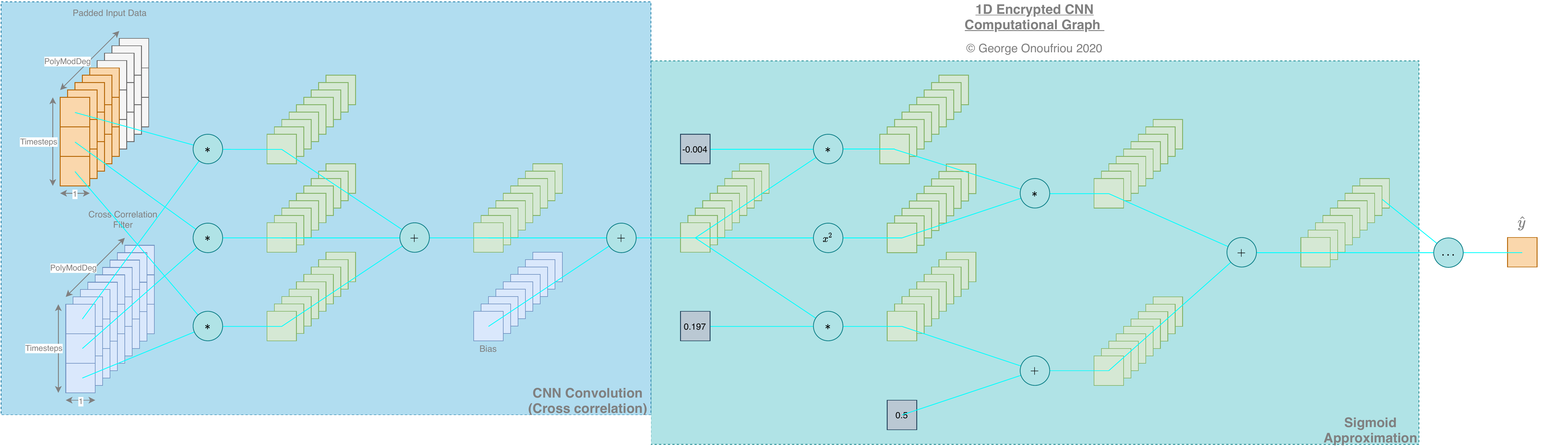}
  \caption{The computational graph of our encrypted 1D CNN, displaying the computational steps and gates necessary to calculate a given output y from input x. This also abstracts these computational steps into groups or modules such as the sigmoid approximation module.}
  \label{fig:backprop}
\end{figure}

Deep learning can broadly be abstracted into three stages, namely: Forward pass, backward pass, and the weight update; Forward pass propagates some input against the weights and internal activation functions of the whole network to produce some form of prediction thus from left to right in the computational graph \ref{fig:backprop}. Backward pass calculates the effect of all weights and biases on the final result by differentiation and the chain rule thus operating from right to left in the computational graph \ref{fig:backprop}. The weight update takes these weights and adjusts them given the loss (a measure of wrongness), and the gradient of the weight in question to approach a lower loss, usually using the gradient descent algorithm.

Fully homomorphic encryption requires that certain constraints be maintained such as the inability to compute division, thus we describe our process as well as how we overcome these obstacles as follows.

\subsubsection{Forward Pass}

As briefly described previously, the forward pass takes some input x, applies some transformation according to the internal weights of the neural network and outputs some prediction (\ref{fig:backprop}). However, depending on the neural network in question, these transformations are usually not FHE compatible, or are not performant under FHE. An example of FHE incompatibility is most activation functions, e.g ReLU, and sigmoid, which require operations such as max and division that are impossible, requiring context and a non abelian operation respectively. To overcome this, we found in literature approximations for  -- in particular -- sigmoid \ref{eq:sigmoid_approx}, which uses polynomials to overcome the barrier of the divisions in the standard sigmoid \ref{eq:sigmoid}:

\begin{equation}
    \sigma(x) = \frac{1}{1+e^{-x}}
    \label{eq:sigmoid}
\end{equation}
where:
\begin{conditions*}
  \sigma & sigmoid \\
  x & some input vector x \\
  e & eulers number
\end{conditions*}

\begin{equation}
    \sigma(x) \approx 0.5 + 0.197x + -0.004x^3
    \label{eq:sigmoid_approx}
\end{equation}
where:
\begin{conditions*}
  \sigma & sigmoid \\
  x & some input vector x \\
\end{conditions*}

This approximation closely follows the standard sigmoid between the ranges of -5 and 5, which is more than sufficient for our purposes since, when normalised most data, weights, and subsequently activations will likely fall in the range 0-1.

This approximation in question, proposed by Chen (\autocite{cryptoeprint:2018:462}) \ref{eq:sigmoid_approx}, is used interchangeably with sigmoid in our equations, thus our neural network equation \ref{eq:neural_network} will stay relatively normal aside from the use of time t as a 1D (time series) CNN.

\begin{equation}
  a^{<t>} = \sigma(w_i^{<t>}x^{<t>}+b_i^{<t>})
    \label{eq:neural_network}
\end{equation}
where:
\begin{conditions*}
  \sigma & sigmoid/ sigmoid approximation \\
  x & some input vector x \\
  e & eulers number
\end{conditions*}

\begin{figure}[th!]
  \centering
  \includegraphics[width=\columnwidth]{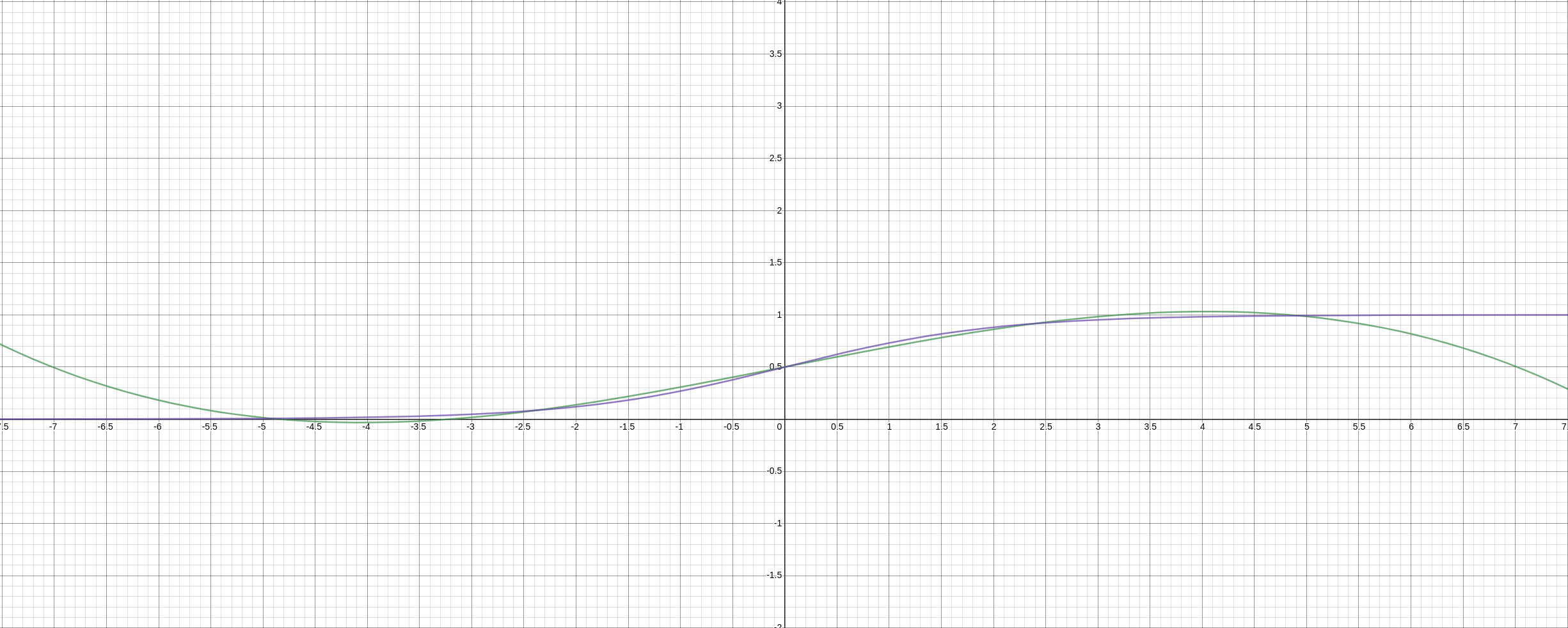}
  \caption{Graphical comparison of sigmoid (purple) and sigmoid approximation (green) functions, showing their similarity between the range of -5 and 5.}
  \label{fig:desmos}
\end{figure}

The specific neural network we used for this forward pass was a 1D convolutional neural network (1D CNN), where we substituted space for time, thus our activation equation becomes what is depicted by equation \ref{eq:neural_network}. We used a 1D CNN over traditional time series neural networks despite having a time series dataset as not only have 1D CNNs been shown to be good time series predictors that are more parrelelisable, but the nature of a CNN means computations are wider rather than deeper. In practice, this means that less expensive operations such as bootstrapping are necessary, since after a few computations deep it is necessary to bootstrap and shrink the ciphertext, thus improving the overall time efficiency of the resulting computational graph. Our CNN is of shape (timesteps, 1, polynomial modulus degree / 2) since the process of encryption in the CKKS scheme utilises a set number of slots based on the polynomial modulus degree used. This means, with the exception of the total length of the encrypted vector, the absolute number of slots populated can not be assessed, and any unpopulated slots are padded with 0s prior to encryption.

 \subsubsection{Backward Pass}
 \label{subsubsection:backward_pass}

 The backward pass, as touched on before, is the process of calculating the gradients of the weights and biases with respect to the output. However to do this you require some cached input which is used to derive the differentials. An example of this dependency of input could be sigmoid (\ref{eq:sigmoid}) which to calculate the gradient requires x the encrypted input (\ref{eq:sigmoid_differential}).

\begin{equation}
  \frac{df}{d\sigma} = (1 - \sigma x) * \sigma x
    \label{eq:sigmoid_differential}
\end{equation}
where:
\begin{conditions*}
  \sigma & sigmoid / sigmoid approximation \\
  x & some input vector x \\
  \frac{df}{d\sigma} & Differential with respect to sigmoid
\end{conditions*}

 Since, to calculate the sigmoids gradient, x (\ref{eq:sigmoid_differential}) is required, this means that either this gradient becomes encrypted in the process of using the currently encrypted x, or that this can only be calculated in plaintext. For now, we choose to calculate gradients in plaintext, as encrypting the gradients would inevitably mean encrypting the new weights, which would mean every operation would become between two ciphertexts, adding substantial time complexity (an order of magnitude), when unnecessary for our purposes, and would function to limit the neural network to that one specific key from that one specific data owner. While this may be desirable in certain bespoke situations, generally, as a data processor however, this is largely undesirable since it is necessary to serve multiple data owners/ sources simultaneously. It may be instead possible to have generic models that can then be privately tailored to the specific data in question.
 For these bespoke models this could potentially be represented somewhat like a graph/ digraph for each individual weight with respect to the generic models, such that if in future the data owner allows these weights to be decrypted, that they can be retroactively used to update other pre-existing models stemming from the same generic model/ node.

 \subsubsection{Weight Update}

 The weight update using gradient descent simply moves the weights in whichever direction approaches a lower/ minimum loss. They do this according to some optimisation function. Equation \ref{eq:weight_update} represents a simple optimisation function we used to test weight update functionality with FHE to approach a lower loss.

\begin{equation}
  w_i^{j+1<t>} = w_i^{j<t>} - (l \frac{df}{dw_i^{j<t>}})
    \label{eq:weight_update}
\end{equation}
where:
\begin{conditions*}
  t & timestep t \\
  l & learning rate \\
  j & weight update iteration \\
  w_i^{j<t>} & some i'th weight at time t with respect to the predicted output iteration j (the current weight update), and j+1 the next weight update \\
\end{conditions*}

%% file: src/sections/results.tex
\begin{table}[h]
  \begin{tabular}{l|ll}
  Operation               & \begin{tabular}[c]{@{}l@{}}Locally \\ (seconds 3.s.f)\end{tabular} & \begin{tabular}[c]{@{}l@{}}Remotely\\ (seconds 3.s.f)\end{tabular} \\ \hline
  Encryption              & 0.0136                                                             & 0.454                                                              \\
  Decryption              & 0.0330*                                                            & 1.14*                                                              \\
  Inference               & 0.966**                                                            & 3.13** *                                                           \\
  Ciphertext + Ciphertext & 0.287                                                              & -                                                                  \\
  Ciphertext + Plaintext  & 0.0480*                                                            & -                                                                  \\
  Ciphertext * Ciphertext & 0.277*                                                             & -                                                                  \\
  Ciphertext * Plaintext  & 0.0500*                                                            & -
  \end{tabular}
  \caption{\label{tab:results}Time complexity of different operations, both locally, and remotely.}
\end{table}

\begin{table}[]
\begin{tabular}{l|l|l|l}
Length & \begin{tabular}[c]{@{}l@{}}Polynomial\\ Modulus Degree\end{tabular} & \begin{tabular}[c]{@{}l@{}}Numpy Plaintext \\ Size (bytes)\end{tabular} & \begin{tabular}[c]{@{}l@{}}Encrypted Vector\\ Size (bytes)\end{tabular} \\ \hline
4096   & 8192                                                                &              32880                                                      &    4800310                                                              \\
8192  & 16384                                                               &              65648                                                      & 9600592
\end{tabular}
  \caption{\label{tab:space}Space complexity of different length vectors, unencrypted as numpy arrays and encrypted as ReSeal vectors, including private keys and all meta-data required for operation.}
\end{table}

In the work presented in this paper, we created a two part client and server system to facilitate EDLaaS at scale, just like any other platform as a service. We also created our own libraries which have consequences on the computational speed of the whole pipeline. Table \ref{tab:results} represents the computational complexity we achieved against an arbitrary dataset, in our case milk yield prediction using a 1D convolutional neural network  as outlined in \ref{subsec:encrypted_deep_learning}. These results are averages of examples and their time of execution, including in the case of the remote examples the transmission time. For the sake of consistency, all results presented were obtained on the same local area network (LAN) to prevent the effect of otherwise uncontrollable conditions and traffic over the wider area network (WAN).
While this is still a fairly small scale relative to production settings, our use of containers can easily be expanded upon such as with kubernetes, apache mesos, or docker swarm, allowing for scaling up and down. However the most difficult task of handling these requests in a scalable, encrypted manner has been tackled, paving the way for such expansions.

It may also be noted in table \ref{tab:results} that some fields are left blank. These blank fields in the remote column are remote operations left unimplemented as they are too low level operations to be worth the overhead cost of transmission, also considering that simple operations alone, such as ciphertext + ciphertext, are not in of themselves a form of deep learning, and would not have been worthy candidates to implement, taking time away from more critical research.

It can be seen in \ref{tab:results} that local time complexity effectively represents the efficiency of our rebinding and abstraction implementation of MS-SEAL, whereas the remote results represent the time complexity of the pipeline overall. That is to say remote encryption, remote decryption, and remote inference. One thing not shown here is time complexity of training, as this is an ongoing area of work for us, where significant decisions must be made about the neural networks as described in \ref{subsubsection:backward_pass}. Along with how best to deal with and integrate bespoke models for problems where the client/ data owner must maintain absolute, uncompromising privacy, thus not allowing any backpropagation to occur.

Space complexity as shown in table \ref{tab:space} however shows how much larger an encrypted cyphertext is relative to its unencrypted counterpart. in the case where the polynomial modulus degree is 16384 and thus the length of the array is half of the poly-mod-degree at 8192, the plaintext numpy array is only 0.0656MB, compared to 9.60MB if it was to be encrypted. However this is the total size of the cyphertext, including all the other required information necessary to store with it, namely its private key, and at the highest point in the modulus switching chain, which is only necessary before computation begins. Thus there will be some gains once unnecessary information to the data processor is stripped, and the data has been computed thus approaching the end of its modulus switching chain, which produces a smaller cyphertext as a side effect, while also being the result of several other cyphertexts combined into a single prediction.
This space complexity is likely easily optimised as this size is a result of some difficult serialisation logic necessary when going from MS-SEAL in C++ to ReSeal in python, and is likely an area where easy gains can be garnered in future.

%% file: src/sections/conclusion.tex
FHE is possibly one of the technologies that in the era of privacy will receive even further attention in the next few years, in particular as a component of machine learning applications. In this paper, We have found that FHE can be successfully applied to deep learning at scale to create Encrypted Deep Learning as a Service. We have also found the time complexity increase to be within acceptable bounds already, despite there being many areas where improvement can be had. In addition, We have conceived and implemented an open-source collection of software to facilitate EDLaaS, which we continue to improve upon. During our developments, we have overcome a plethora of difficulties with regards to combining deep learning with FHE, which we have discussed here in detail along with our solutions/ mitigations. Finally, we have outlined a few areas where special consideration is required in the future, such as bespoke models with encrypted weights.